%% file: main.tex
\documentclass[letterpaper]{article} 
\usepackage[preprint]{aaai2027} 
\usepackage[hyphens]{url} 
\usepackage{graphicx} 
\newcommand{\tbthreefigroot}{figures}
\IfFileExists{figures/adjudication_pipeline.pdf}{}{\renewcommand{\tbthreefigroot}{aaai-draft/figures}}
\usepackage{natbib} 
\usepackage{caption} 
\usepackage{booktabs}
\usepackage{amsmath}
\title{What Makes a Terminal-Bench Task Hard?\\
Separating Genuine Hardness from Fake-Hardness on an Adjudicated Agentic Corpus}

\author{
    Edward Lue Chee Lip,
    Boden Moraski,
    Tim Knappe,
    Lang Xiong,
    Sarvesh Gharat,
    Antonio Mari,
    Ivan Bercovich
}
\affiliations{
    [Affiliations and contact e-mail to be added]
}

\begin{document}
\maketitle
\begin{abstract}
\input{sections/00-abstract}
\end{abstract}

\input{sections/01-introduction}
\input{sections/02-corpus-method}
\input{sections/03-certified-frontier}
\input{sections/04-rejection-taxonomy}
\input{sections/05-cost}
\input{sections/06-discussion-limitations}
\input{sections/07-conclusion}
\label{main-end}

\bibliography{refs}
\end{document}

%% file: sections/00-abstract.tex
Frontier benchmarks need tasks that current models cannot solve. But a task that no model solves is not automatically a hard task. The same zero pass rate can come from a real capability gap, but it can also come from missing context, a broken reference solution, infrastructure failure, or a verifier that can be bypassed. In this paper, we study this issue using a frozen Terminal-Bench~3 / Frontier-Bench~0.1 production record with 1,081 pull requests, 639 scored tasks, 28,801 trials, and \$105,933 in logged agent spend. We ask what an all-fail task actually certifies. For the 125 tasks with no honest pass, we combine task artifacts, reference-solution runs, empty-solution controls, adversarial trials, trajectories, telemetry, and review records, and apply an ordered validity screen. Only 78 of the 125 tasks survive as certified-unsolved candidates. The remaining tasks include 14 with broken oracles, 8 dominated by infrastructure failures, 4 that are only passable through verifier bypasses, and 21 whose solvability is not certified by the available evidence. Thus, lack of saturation and genuine difficulty are not the same thing. The certified-unsolved label is also narrow: it means that the authored route passed, infrastructure did not dominate, no strict bypass was observed, and all evaluated agents failed. It does not prove intrinsic hardness, verifier completeness, or failure at the intended capability. We further analyze rejected submissions and passing tasks to show that pass rate alone cannot explain why a task is difficult. Overall, our results suggest that frontier benchmarks should report the evidence behind their all-fail tasks before using them as capability claims.

%% file: sections/01-introduction.tex
\section{Introduction}
\label{sec:intro}

A frontier benchmark is useful only if it leaves room for progress. If current models solve nearly every task at release, the benchmark cannot say much about the next generation of systems. This makes low pass rate an important design goal, but not a sufficient validity guarantee. A task may have zero passes because it requires a capability current agents do not have. But the same outcome can also occur when the task is underspecified, the reference solution is broken, the environment is unreliable, or the verifier accepts the wrong behavior. In all these cases, the leaderboard reports the same number, even though the scientific meaning of that number is different.

This distinction is important because benchmarks do more than rank models. They shape what the field treats as progress. When a task is accepted as a frontier task, it becomes evidence about what current systems cannot do and what future systems should optimize for. If all-fail tasks are accepted without checking why they failed, benchmark construction can reward brittle or poorly specified tasks rather than tasks that expose genuine capability gaps.

The problem is especially visible in agentic benchmarks. Unlike static question-answer benchmarks, an agentic task contains several moving parts: a natural-language instruction, files, dependencies, tools, an execution environment, a reference solution, and a verifier. A failure can arise at any one of these layers. Thus, pass rate alone cannot determine whether the agent failed at the intended capability or whether the task failed as a measurement instrument.

In this paper, we study this issue using a frozen Terminal-Bench~3 production record, also referred to as Frontier-Bench~0.1. The snapshot contains 1,081 pull requests, 639 scored tasks, 28,801 trials, and \$105,933 in logged agent spend. We treat this record as a task-production corpus rather than only as a final leaderboard. This is useful because the archive contains not only accepted tasks, but also pull-request discussions, reference-solution runs, empty-solution controls, adversarial trials, execution logs, trajectories, and rejected submissions. These artifacts let us ask what evidence supports each task-level failure.

Our main focus is the set of 125 tasks with no honest pass. For these tasks, pass rate gives no variation: every task appears equally unsolved. We therefore apply an ordered validity screen. The screen asks whether the authored reference route passes, whether infrastructure failures dominate the observed attempts, whether there is reward-independent evidence of a verifier bypass, and whether the remaining evidence is sufficient to certify one working route. This produces five mutually exclusive outcomes. Of the 125 all-fail tasks, 78 survive as certified-unsolved candidates. The remaining 47 do not support the same interpretation: 14 have broken oracles, 8 are infrastructure-limited, 4 are exploit-only-passable, and 21 have uncertified solvability.

The certified-unsolved label is deliberately limited. It means that, in the frozen setup, the authored route passed, infrastructure did not dominate, no strict bypass evidence was observed, and all evaluated agents still failed. It does not prove intrinsic hardness, verifier completeness, or that every failed agent reached the intended conceptual crux. The purpose of the label is to state what the available evidence supports, while keeping unresolved cases separate from genuine hardness claims.

We also use the production record to study how misleading difficulty enters benchmark construction. The 555 closed-unmerged submissions show that rejected tasks are not simply easy tasks. Many are tasks that agents fail, but that review rejects for reasons such as ambiguity, verifier problems, nondeterminism, metadata issues, duplication, or insufficient task quality. This shows why low pass rate is useful as a coarse signal but insufficient as a diagnosis. A task can be hard for the wrong reason.

Finally, we study cost as a signal beyond pass rate. Among tasks with at least one honest pass, the minimum output tokens needed to obtain the first archived pass provide a graded measure of effort. This measure separates broader genuine-hard candidates from tractable tasks with AUC 0.750. This result is observational and does not define intrinsic hardness, but it shows that useful difficulty information remains after the binary solved/unsolved view has saturated.

Our contributions are as follows.
\begin{itemize}
\item We reconstruct and audit a frozen Terminal-Bench~3 / Frontier-Bench~0.1 production corpus containing 1,081 pull requests, 639 scored tasks, 28,801 trials, reference and empty-solution controls, adversarial trials, trajectories, telemetry, and review records.
\item We provide an item-validity analysis of the 125 tasks with no honest pass. The analysis separates a single all-fail leaderboard value into five evidence-qualified outcomes: 78 certified-unsolved candidates, 14 broken-oracle tasks, 8 infrastructure-limited tasks, 4 exploit-only-passable tasks, and 21 tasks with uncertified solvability.
\item We analyze 555 closed-unmerged submissions to characterize where misleading difficulty enters task production. The rejection record shows that many failed candidate tasks are not valid frontier items, even when agents struggle with them.
\item We study effort beyond pass rate using the minimum output tokens to first honest pass. Among passing tasks, this cost-based signal gives a graded measure of difficulty and separates broader genuine-hard candidates from tractable tasks with AUC 0.750.
\end{itemize}

Overall, this paper is not a model ranking. It is an audit of what an all-fail benchmark result can certify. Our results suggest that frontier benchmarks should report the evidence behind all-fail tasks before using them as capability claims: a working reference route, a failing empty control, reachable infrastructure, verifier-integrity evidence, declared harness conditions, and an explicit record of unresolved uncertainty.

These results turn an all-fail score into a more useful benchmark signal. They identify which tasks can be retained as frontier candidates, which tasks need targeted repair, and which cases should remain unresolved until more evidence is available. This makes the benchmark easier to trust and easier to maintain: model failures become more interpretable, task-production defects become visible, and future releases can state not only how often agents fail, but also what those failures support.

\paragraph{Related work.}
This work relates to several lines of benchmark evaluation. Item-response-theoretic audits estimate item difficulty, feasibility, and guessing behavior from model response patterns \citep{zhou2025lost,land2026auditing}, and recent work extends this view to agentic coding benchmarks \citep{ge2026agentpsych}. These methods are closely related to our difficulty--feasibility split, but they infer item properties from the same responses used to compute the score. In contrast, we provide adjudicated item-validity labels from production evidence, including task artifacts, controls, trajectories, telemetry, and review records. Work on agent harnesses shows that scaffolds and execution setups can materially change benchmark outcomes \citep{yao2026harnessbench,gorinova2026position}. We therefore treat every score as conditional on the recorded task, harness, verifier, and environment, rather than making context-free model comparisons. Verifier hardening, reward-independent audits, and reward-hacking studies show that weak checkers can be exploited and, in training settings, can shape model behavior \citep{zhong2026hardening,wang2026benchjack,helff2026gaming,bercovich2026terminalwrench}. In our analysis, verifier bypass evidence is one part of the adjudication protocol, not the whole object of study. Similarly, calibrated LLM verifiers can improve solution checking \citep{kwok2026llmverifier}, but a better verifier still scores the task as authored; it does not by itself establish that the task supports the intended capability claim. Contamination audits \citep{song2026crosscontext,prathifkumar2025swebench} and benchmark-reporting checklists \citep{zhu2025bestpractices,kim2026fivenines} address complementary evaluation-hygiene problems. Our contribution is to join these concerns at the item level for a deployed agentic benchmark, and to ask what each all-fail task actually supports.

%% file: sections/02-corpus-method.tex
\section{Corpus and Adjudication Protocol}
\label{sec:method}

We use the frozen Terminal-Bench~3 / Frontier-Bench~0.1 production record as a task-production corpus, not only as a final leaderboard. This is important because an all-fail score cannot be interpreted from the score alone. We need to know what task was submitted, what was executed, what the controls showed, and whether the observed failures came from the intended problem or from another layer of the evaluation stack.

All quantitative results use the frozen signal-table snapshot \texttt{3c5be84efd707da8}. The snapshot contains 1,081 pull requests, 639 scored tasks, 28,801 trials, and \$105,933 in logged agent spend. The trials include 21,219 ordinary agent trials, 4,668 cheat-variant trials, 1,605 oracle runs, and 1,309 nop runs. The PR corpus is frozen through PR 1416, so later changes to the live benchmark do not affect the reported results. Each task is linked to its submitted package, pull-request history, verifier, reference solution, trial telemetry, trajectories, controls, and review records.

\begin{figure*}[t]
\centering
\includegraphics[width=\textwidth]{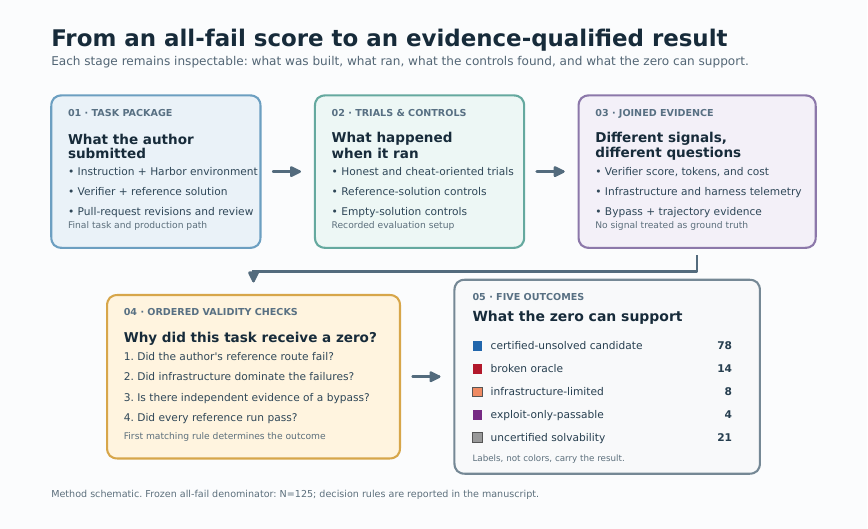}
\caption{Overview of the adjudication protocol. The audit starts from the submitted task package, joins ordinary trials with reference-solution, empty-solution, and cheat-oriented controls, and then combines these runs with telemetry, trajectory evidence, and verifier-integrity evidence. The ordered validity checks assign an evidence-qualified outcome to each task with no honest pass. The outcome counts are shown for orientation; Section~\ref{sec:allfail} analyzes them.}
\label{fig:pipeline}
\end{figure*}

Figure~\ref{fig:pipeline} summarizes the adjudication flow. The first stage records what the author submitted, including the instruction, environment, verifier, reference solution, and review history. The second stage records what happened when the task was run, including ordinary trials, cheat-oriented trials, reference-solution controls, and empty-solution controls. The third stage joins these runs with telemetry, token and cost data, trajectory evidence, and reward-independent bypass evidence. No single signal is treated as ground truth. Instead, the joined evidence is passed through an ordered validity screen that assigns one of five outcomes to each task with no honest pass.

We separate observed hardness into two coordinates. \emph{Difficulty} is the realized effort of solving a task in the recorded setup, measured descriptively using output-token cost. \emph{Feasibility} asks whether a correct solution is reachable and accepted by the task as shipped. This distinction is necessary because pass rate mixes several failure modes. If an ordinary agent passes, then the task has at least one witnessed feasible route. If no ordinary agent passes, feasibility cannot be inferred from the zero and must be adjudicated using the available controls and task evidence.

The controls answer different questions. An \emph{oracle run} executes the author's reference solution and tests whether the intended route is accepted by the shipped verifier. A \emph{nop run} submits an empty solution and tests whether the verifier rejects trivial non-solutions. A \emph{cheat-variant trial} prompts an agent to pass by any available route and probes what the verifier can be induced to accept. A passing oracle establishes one accepted route in the recorded setup, but it does not prove verifier completeness. A failing nop control rules out the simplest empty-solution leak, but it does not rule out all verifier bypasses.

Our main adjudication applies to tasks with no honest pass. An honest pass is a passing ordinary agent trial, excluding oracle, nop, and cheat-variant runs. For each all-fail task, we apply a first-matching rule. If the authored reference route fails, the task is labeled \emph{broken oracle}. If infrastructure failures account for at least half of ordinary failures, it is labeled \emph{infrastructure-limited}. If there is strict reward-independent evidence that the task is passable through verifier bypass, it is labeled \emph{exploit-only-passable}. If at least one oracle run is observed and every recorded oracle run passes, the task is labeled a \emph{certified-unsolved candidate}. The remaining cases are labeled \emph{uncertified solvability}.

Strict bypass evidence requires a persisted artifact, recovered independently of the reward, that defeats the intended check at verification time. A suspicious score gap, access to verifier files, or a transient modification that the agent later reverts is not enough. The certified-unsolved label is therefore narrow. It means that one authored route passed, infrastructure did not dominate the failures, no strict bypass was found, and all evaluated ordinary agents failed in the frozen setup. It does not prove intrinsic hardness, verifier completeness, or failure at the intended conceptual crux.

%% file: sections/03-certified-frontier.tex
\section{What an All-Fail Score Can Support}
\label{sec:allfail}

\begin{figure}[t]
\centering
\includegraphics[width=\columnwidth]{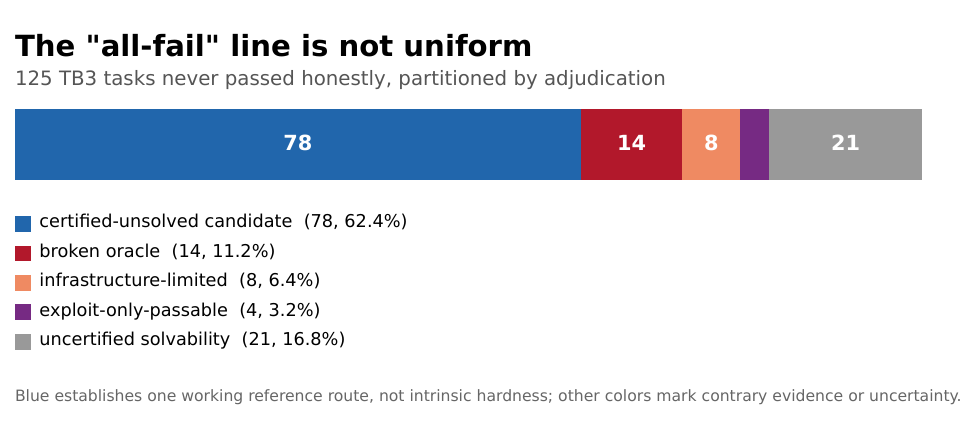}
\caption{One leaderboard value produces five evidence-qualified outcomes. Of the 125 tasks with no honest pass, 78 retain a working reference route and survive the declared validity screens, while 47 require repair, exclusion, or an unresolved label. Counts are computed from snapshot \texttt{3c5be84efd707da8}.}
\label{fig:decomp}
\end{figure}

We now apply the adjudication protocol to the tasks with no honest pass. These are the tasks where the leaderboard gives the least information, because every task receives the same observed outcome. Without additional evidence, a genuinely difficult task, a broken task, and an unresolved task all appear identical.

Figure~\ref{fig:decomp} shows the resulting partition. Among the 125 tasks with no honest pass, 78 survive as certified-unsolved candidates. The remaining 47 do not support the same interpretation. Fourteen have a broken oracle, 8 are infrastructure-limited, 4 are exploit-only-passable, and 21 have uncertified solvability. Thus, an all-fail score is not a single kind of evidence. It can support retention, repair, exclusion, or an unresolved label, depending on which layer produced the zero.

The five outcomes also give different maintainer actions. A certified-unsolved candidate can be retained as frontier evidence under the recorded checks. A broken-oracle task needs repair of the reference route, verifier, or task package before its zero is meaningful. An infrastructure-limited task should not be used as a capability claim until the environment is reachable. An exploit-only-passable task should be excluded or redesigned because the observed passability comes through the verifier rather than the task. An uncertified task should remain unresolved until more evidence is available.

The distinction is visible at the task level. \texttt{iron-codex} has no honest pass across 33 trials, but the authored reference route passes and infrastructure does not dominate the failures. Its failed attempts also consume a large amount of output, with median failed-trial cost of 275,373 tokens. Under the frozen evidence, this task remains a certified-unsolved candidate. In contrast, \texttt{pipeline-tied-weights} is also all-fail, but all 18 ordinary attempts end in infrastructure failure before substantive work. Its zero therefore does not support the same conclusion. Both tasks print as 0\%, but only one currently supports a frontier-task interpretation.

\begin{table}[t]
\centering
\small
\begin{tabular}{lr}
\toprule
\textbf{Certification requirement} & \textbf{Tasks} \\
\midrule
Never passed honestly & 125 \\
Observed oracle pass & 85 \\
Also infrastructure-clean at $\tau = 0.5$ & 79 \\
Also no strict bypass evidence & 78 \\
Also at least two oracle runs & 25 \\
\bottomrule
\end{tabular}
\caption{Certification strictness ladder for the all-fail set. Each row adds one requirement to the preceding row. The final row is a stronger stability cut, not evidence from an independent implementation.}
\label{tab:ladder}
\end{table}

The certified count should be read through the strictness ladder in Table~\ref{tab:ladder}. Certification is not a claim of impossibility. It is the fourth rung of a sequence of checks. Of the 125 never-passed tasks, 85 have an observed oracle pass. After requiring infrastructure to be clean at the default threshold, 79 remain. After also requiring no strict bypass evidence, 78 remain certified-unsolved candidates. If we further require at least two recorded oracle runs, the count falls to 25.

This ladder is important because the corpus does not provide the same amount of evidence for every task. Fifty-three of the 78 certified-unsolved candidates have only one recorded reference run. A single passing reference route shows that one authored solution is accepted once in the recorded setup. It does not show that the verifier is complete, that all valid routes would pass, or that the ordinary agents failed at the intended conceptual crux. Repeated reference runs and independent implementations would strengthen the claim.

\paragraph{Reliability and sensitivity of the adjudication.}
Because the main result depends on task-level labels, we also check how much support the archive gives to the adjudication itself. These checks answer different questions and should not be read as a single accuracy estimate. They are meant to show where the labels are stable, where they are only evidence-qualified, and which cases need stronger review.

The table also clarifies why we keep the certified-unsolved label narrow. A certified task has one accepted reference route and no visible infrastructure or strict-bypass disqualifier under the frozen protocol. This is stronger than a raw all-fail score, but weaker than a proof of intrinsic hardness. In particular, only 25 of the 78 certified-unsolved candidates have at least two recorded oracle runs. The larger set is therefore appropriate for the main audit, while the repeated-oracle subset is a useful higher-confidence cut.

The infrastructure threshold is similarly not the main driver of the result. The default rule labels a task infrastructure-limited when infrastructure failures account for at least half of ordinary failures. Re-running the same ordered rule at nearby thresholds changes only a small number of task IDs. This does not make the threshold unique, but it shows that the headline conclusion is stable: the all-fail set contains a large certified group and a substantial set of tasks that require repair, exclusion, or unresolved status.

\begin{table}[t]
\centering
\small
\begin{tabular}{@{}p{0.34\columnwidth}p{0.17\columnwidth}p{0.38\columnwidth}@{}}
\toprule
\textbf{Check} & \textbf{Result} & \textbf{Interpretation} \\
\midrule
Judge-panel unanimity on contested trials & 48/53 & Most contested trial-level decisions were stable across repeated judgments. \\
\addlinespace
Panel-majority agreement with hand labels & 41/53 & The automatic panel gives useful but imperfect agreement with human labels. \\
\addlinespace
Adversarial re-check & 12/12 held & Selected labels survived a targeted attempt to overturn them. \\
\addlinespace
Infrastructure-threshold sensitivity & 1--3 task changes & The five-way partition is not driven by a single fragile threshold choice. \\
\addlinespace
Repeated-oracle cut & 25/78 & A stricter certified subset remains when repeated reference evidence is required. \\
\bottomrule
\end{tabular}
\caption{Reliability and sensitivity checks for the adjudication. Each row tests a different source of uncertainty. These checks support the evidence-qualified labels, but they do not imply that every task received full domain-expert re-review.}
\label{tab:reliability}
\end{table}

Taken together, these checks make the interpretation more transparent. The labels are not presented as omniscient truth. They are archived, auditable decisions supported by controls, trajectories, telemetry, and review evidence. This is the level of evidence needed for the paper's claim: not that the certified tasks are impossible, but that a benchmark zero should be interpreted only after the evidence behind it is made visible.

The validity screen and the depth question should therefore be kept separate. The screen asks whether the task can be completed and graded in the frozen setup. The depth question asks what the failed attempts actually exercised. These answers can differ. A task may have a passing reference route, while ordinary agents spend most of their time on setup, dependency resolution, or other friction around the intended problem. The separate depth audit leaves such cases unresolved when the trajectory evidence does not identify the failure layer.

The main result of this section is that the all-fail set becomes more useful after adjudication. The 78 certified-unsolved candidates are stronger evidence for the benchmark frontier than the raw zero alone. At the same time, the 47 remaining tasks are not wasted information. They show where the benchmark needs repair, exclusion, or additional evidence. The next section asks where such misleading difficulty enters the task-production process by studying the rejected submissions.

%% file: sections/04-rejection-taxonomy.tex
\section{How Misleading Difficulty Enters Task Production}
\label{sec:rejected}

\begin{figure*}[t]
\centering
\includegraphics[scale=0.7]{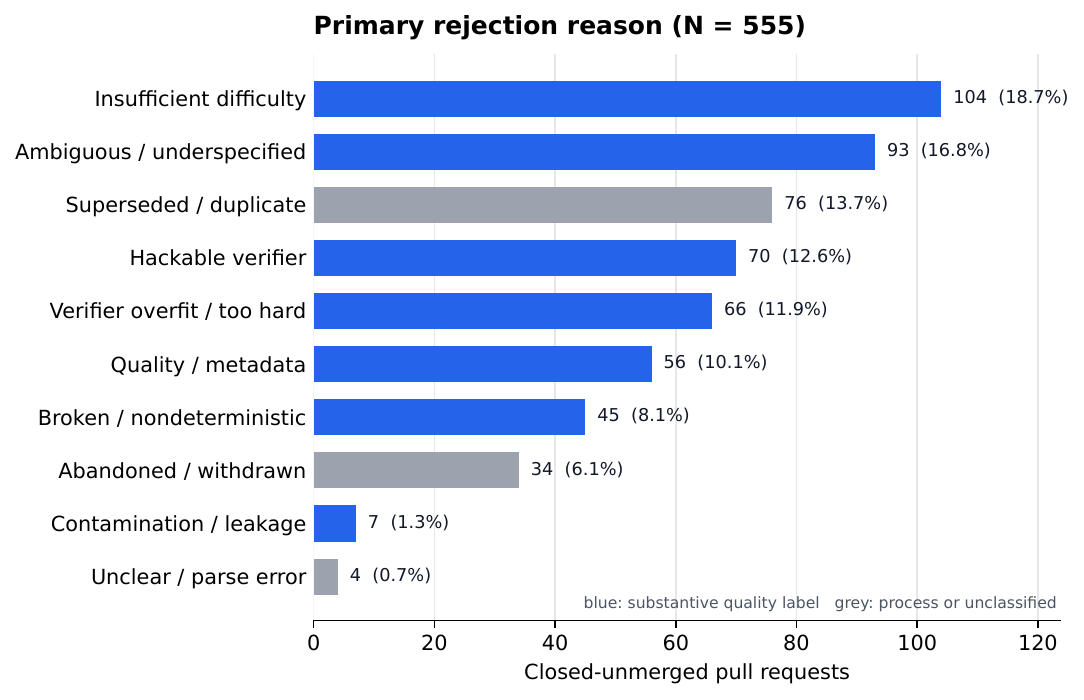}
\caption{Primary rejection reason for the 555 closed-unmerged Terminal-Bench~3 pull requests. The chart shows that rejected tasks are not only tasks that agents solve too easily. Many are rejected because of ambiguity, verifier problems, nondeterminism, metadata issues, duplication, or process closure. Labels are assigned from review threads by a single-pass LLM judge.}
\label{fig:rejection}
\end{figure*}

The previous section studies tasks after they enter the scored benchmark. The production record also lets us look earlier, at candidate tasks that were proposed but did not merge. This matters because low pass rate can appear before a task is valid. A candidate task may defeat agents because it is genuinely demanding, but it may also defeat them because the instruction is unclear, the verifier is too brittle, the environment is nondeterministic, or the task is missing information needed to make the intended solution reachable.

We classify the 555 closed-unmerged pull requests by their primary rejection reason. Figure~\ref{fig:rejection} shows the resulting distribution. The largest category is insufficient difficulty, but many other rejected tasks fail for reasons that are closer to task validity than to model capability. These include ambiguous or underspecified instructions, hackable verifiers, verifier overfit, broken or nondeterministic setups, and quality or metadata issues. Thus, the rejection record shows that benchmark construction is not only a search for harder tasks. It is also a filtering process that removes tasks whose apparent difficulty comes from the wrong source.

Pass rate is still useful, but only as a coarse signal. It separates some extremes. Tasks rejected for insufficient difficulty have mean honest pass rate $0.64$, while tasks rejected as verifier-overfit or too hard have mean pass rate $0.10$. This agrees with the intuition that pass rate contains information about difficulty. The problem is that it does not diagnose the reason for difficulty. Ambiguous or underspecified tasks have mean pass rate $0.17$, and broken or nondeterministic tasks have mean pass rate $0.18$. These values lie in the same low-pass region as tasks that may be genuinely difficult. A leaderboard score alone therefore cannot tell whether a task is unclear, unsatisfiable, broken, over-constrained, or simply hard.

Some defects are even less visible from pass rate. Hackable-verifier rejections have mean pass rate $0.41$, and quality or metadata rejections have mean pass rate $0.45$. These are not necessarily low-pass tasks, but they are still invalid or unsuitable benchmark items. This is another reason why pass rate should not be used as the only acceptance criterion. It can miss defects that do not strongly affect solve rate, and it can conflate very different defects when solve rate is low.

The rejected submissions are also not just trivial tasks. Among the 349 rejected pull requests whose task reached the trial harness, only about $31\%$ have honest pass rate above one half. About $42\%$ have pass rate at most $0.2$, and the rest fall in the mixed region. After excluding tasks with clearly defective verifier evidence, the failed-by-agents share increases to about $45\%$. Thus, many rejected tasks are tasks that agents struggle with. They are rejected because the struggle does not cleanly support a frontier capability claim.

This is where production history becomes useful. Pull-request revisions, reviewer objections, control outcomes, and trial failures show which part of the task-production process caught a defect. A missing file, an ambiguous instruction, a verifier coupled to one route, and a fragile environment leave different traces in the archive. Preserving those traces makes the benchmark easier to maintain, because future task authors can see which checks failed and why.

We use provenance in this operational sense, not as a ranking of contributors or sources. The stable lesson is process-level. A benchmark owner should be able to ask which checks catch ambiguity, which checks catch verifier failure, which checks catch infrastructure problems, and which cases remain unresolved. The rejected-task record therefore complements the all-fail analysis. The all-fail screen tells us what a zero can support after a task is accepted. The rejection record shows how many tempting zeros must be filtered out before they become benchmark evidence.

%% file: sections/05-cost.tex
\section{Cost Beyond Pass Rate}
\label{sec:cost}

The all-fail analysis tells us what a zero can support, but it cannot measure the cost of a successful solution. By definition, the all-fail set contains no honest passing trial. We therefore ask a separate question on tasks that do pass at least once: among solvable tasks, does the cheapest recorded success still carry useful information about difficulty?

For each task with at least one honest pass, we compute the minimum output tokens used by any archived honest passing trial. This gives a conservative effort measure, since it records the cheapest observed success rather than the average cost over all attempts. We use output tokens because they are available across the recorded runs and directly reflect the amount of agent work produced before the verifier accepted a solution.

The analysis covers 346 tasks with at least one honest pass. We compare 217 broader genuine-hard candidates with 129 tractable tasks. Both groups require a non-broken oracle and no reward-hack gap; verifier-suspect tasks are excluded. The groups are then split by honest pass rate, while the measured outcome is token cost. Thus, the comparison asks whether tasks that are rarely passed also require more output when they are passed.

They do. The broader genuine-hard candidates have median minimum successful cost of 24,829 output tokens, compared with 12,388 for tractable tasks. The resulting separation has AUC 0.750 with 95\% confidence interval [0.700, 0.801]. In pairwise terms, a randomly chosen broader genuine-hard candidate has a 75\% chance of having a higher cheapest-success cost than a randomly chosen tractable task.

This result should be read as observational. Minimum tokens to first pass is not intrinsic hardness. It mixes reasoning, verbosity, tool use, harness policy, model assignment, retry behavior, budget, and task revision. It also exists only for tasks where a pass was observed. Still, it is useful because it recovers a graded signal in the region where pass rate is too coarse. Two tasks may both be solvable, but one may require much more recorded work before any agent reaches an accepted solution.

The same observation motivates a controlled hint ladder, but does not replace it. A hint experiment would hold the task revision, environment, verifier, model, harness, budget, tools, and retry policy fixed, while adding preregistered information across repeated trials. If a small hint reliably flips a task from failure to success, the original failure may have depended on missing context rather than deeper capability. If success appears only after substantial guidance, the difficulty may be more robust. We do not report such an intervention here; the ladder remains a follow-up experiment requiring owner approval, compute, and domain review.

Overall, token cost complements the validity audit. The audit asks whether a task failure can be interpreted as evidence. The cost signal asks how much recorded effort was needed when success was possible. Together, they give a more useful view than pass rate alone: whether the task is valid enough to measure, and how much work the recorded agents needed to solve it.

%% file: sections/06-discussion-limitations.tex
\section{Discussion and Limitations}
\label{sec:discussion}

The main lesson is that an all-fail score becomes useful only after its source is known. A leaderboard can tell us that no recorded agent passed a task, but it cannot tell us why. The failure may come from the intended capability, from missing context, from the environment, from the harness, or from the verifier. The same score can therefore support different decisions. Some tasks should be retained as frontier candidates, some should be repaired, some should be excluded, and some should remain unresolved until more evidence is available.

This gives a practical reporting standard for frontier benchmarks. A hard task should not be released only with its pass rate. It should also report whether the reference route passes, whether an empty solution fails, whether the required tools and environment are reachable, whether verifier integrity was checked independently of the reward, and whether observed failures reached the claimed crux. These checks do not make the benchmark perfect, but they make the meaning of a failure inspectable. They also make benchmark maintenance easier, because task defects become visible before they are mistaken for capability gaps.

The same point applies to model comparisons. Scores are conditional on the task, environment, verifier, model, harness, provider state, budget, and retry policy. This paper does not try to remove that dependence or produce a context-free ranking. Instead, it makes the dependence explicit. A benchmark can still compare systems, but the comparison is more useful when the instrument that produced the score is pinned and the unresolved cases are reported.

There is also a broader construct question. A task can be technically valid in the sense that it has a working reference route and a sound verifier, while still measuring a capability that is not the most important target for the field. This paper addresses the first issue. It asks whether the shipped item supports the capability claim it makes. It does not decide which capabilities benchmark builders should prioritize. That choice should be stated by benchmark publishers, together with the reason that progress on the chosen construct matters.

The labels in this paper are evidence-qualified rather than final truth. Oracle coverage is uneven, and 53 of the 78 certified-unsolved candidates rely on a single recorded reference run. A passing reference route establishes one accepted solution in the recorded setup. It does not prove verifier completeness, deterministic behavior, or acceptance of alternative valid routes. Repeated reference runs and independently implemented solutions would make the certification stronger.

The trajectory and depth evidence also has limits. Some failures clearly occur at the claimed crux, while others are dominated by setup, dependencies, or incomplete records. We keep such cases unresolved when the archive does not identify the failure layer. This is important because uncertainty should not be converted into hardness. An indeterminate case is not halfway between genuine-hard and fake-hard; it is a case where the available evidence is not enough.

Our model and harness evidence is observational. The recorded runs are not a fully crossed experiment over models, harnesses, budgets, provider states, and task versions. Token cost also mixes several factors, including reasoning, verbosity, tool use, retry behavior, and harness policy. The AUC result in Section~\ref{sec:cost} should therefore be read as a descriptive signal, not a causal estimate of intrinsic difficulty.

The rejection taxonomy has similar limits. Each rejected pull request receives one primary reason, even though several defects can occur together. The labels are based on review records and a single-pass LLM-judge protocol rather than duplicate domain-expert annotation of every thread. They describe the recorded production decision, not a unique causal mechanism. Contributor and provenance metadata are also incomplete, so aggregated provenance should be used for process improvement rather than for ranking task sources.

Finally, all results are tied to one frozen snapshot of an evolving benchmark. Later repairs, new tasks, new model runs, or changed harnesses can alter the raw counts. This is why the snapshot identifier is part of each quantitative claim. The controlled hint ladder and matched harness experiments remain follow-up studies, not results of this paper. Within these limits, the audit still gives a concrete improvement over raw pass rate: it states what each all-fail task can support, what remains unknown, and what evidence would make the claim stronger.

%% file: sections/07-conclusion.tex
\section{Conclusion}
\label{sec:conclusion}

A benchmark task that no agent solves can be useful evidence, but only after we know what produced the failure. In the frozen Terminal-Bench~3 / Frontier-Bench~0.1 record, 125 tasks share the same all-fail score, yet they do not support the same conclusion. Seventy-eight survive as certified-unsolved candidates, while the remaining tasks require repair, exclusion, or more evidence. The certified label is intentionally limited: it records one working reference route and the absence of visible infrastructure and strict-bypass disqualifiers in the frozen setup; it does not prove intrinsic hardness. The broader lesson is that frontier benchmarks should make failures inspectable. Reporting controls, infrastructure status, verifier-integrity evidence, trajectory uncertainty, and task-production history turns a raw zero into a benchmark signal that is easier to trust, maintain, and improve.

\section*{Generative AI Disclosure}
The authors use ChatGPT and Claude for language editing and formatting. They also use the Stanford Agentic Reviewer\footnote{\url{https://paperreview.ai/}} and the CMU Paper Reviewer \citep{kim2026reviewers} to obtain preliminary feedback and improve the paper. All technical content, experimental design, analyses, and conclusions remain the sole responsibility of the authors.